\documentclass{article}

\usepackage{spconf}

\usepackage[utf8]{inputenc} % allow utf-8 input
\usepackage[T1]{fontenc}    % use 8-bit T1 fonts
\usepackage{amsfonts,amsmath,bm}
\usepackage{stmaryrd}
\usepackage{microtype,nicefrac}
\usepackage{booktabs}
\usepackage{graphicx}
\usepackage{hyperref}
\usepackage{balance}

\title{FAST LEARNING FROM LABEL PROPORTIONS WITH SMALL BAGS}

\name{Denis Baručić, Jan Kybic}
\address{Department of Cybernetics, Faculty of Electrical Engineering, Czech Technical University in Prague}

\renewcommand{\vec}[1]{\bm{#1}}

\newcommand{\bag}[1]{\vec{\uppercase{#1}}}
\newcommand{\inst}[1]{\vec{\lowercase{#1}}}

\newcommand{\etal}{\textit{et al}.}
\def\etal.{et\penalty50\ al.}

\def\H{{\mathcal{H}}}
\def\h{{\vec{h}}}

\begin{document}

\maketitle

\begin{abstract}
In learning from label proportions (LLP), the instances are grouped into bags,
and the task is to learn an instance classifier given relative class
proportions in training bags. LLP is useful when obtaining individual instance
labels is impossible or costly.

In this work, we focus on the case of small bags, which allows to design an
algorithm that explicitly considers all consistent instance label combinations.
In particular, we propose an EM algorithm alternating between optimizing a
general neural network instance classifier and incorporating bag-level
annotations. Using two different image datasets, we experimentally compare this
method with an approach based on normal approximation and two existing LLP
methods. The results show that our approach converges faster to a comparable or
better solution. 
\end{abstract}


\section{Introduction}

\textit{Learning from label proportions} (LLP) \cite{musicant2007} is a~case of
a~weakly supervised learning framework where the aim is to produce an instance
classification model, knowing the proportions of class labels in groups of
instances \textit{(bags)} but not the individual instance labels.

The first applications of LLP were in modeling voting behavior from aggregated
electoral district data, where individual data is not available because of
privacy requirements.  In \textit{in vitro} fertilization (IVF), the task is to
predict the likelihood of successful development of individual embryos
\cite{hernandez2018} or oocytes \cite{barucic2021}, while the only hard data is
the outcome of the pregnancy, which may have resulted from any of the several
implanted embryos. In Langerhans islet detection~\cite{Habart2016}, only
subjective classification of individual objects is possible, while the total
contents in the sample can be quantified by DNA content measurement. In
counting applications, i.e., counting the number of people, animals, cells,
cars, or other objects in the image \cite{lempitsky2010}, it is often cheaper
to obtain the counts than annotating individual objects. While evaluating
histopathology samples, instead of painstakingly delineating the cancerous
tissue, experts usually give a~``grade'' corresponding to the extent of this
tissue~\cite{bandi2018detection}.  Similarly, in 3D CT lung volumes, it is much
easier for the expert to quantitatively estimate the extent of emphysema
instead of performing a~complete pixel-wise segmentation~\cite{bortsova2018deep}. 
In remote sensing, a~pixel-wise classifier for SAR images was successfully
trained from label proportions in low-resolution grid
cells~\cite{ding2017learning}.  Other applications of LLP include fraud
detection \cite{rueping2010}, video event detection \cite{lai2014}, or learning
attribute-based representations of images \cite{yu2014}.

\subsection{Related work}

LLP was formulated by Musicant~\cite{musicant2007} and has been approached by
many traditional machine learning techniques~\cite{dulac2019}. Among support
vector machines (SVM) methods \cite{rueping2010,chen2017}, Alter-$\propto$SVM
\cite{yu2013} has received the most attention. It is an example of the
Empirical Proportion Risk Minimization (EPRM) framework \cite{yu2015}, which
learns an instance classifier by minimizing a~bag-level loss function defined
on the label proportions. The {Alter-$\propto$SVM} algorithm is iterative and
alternates between learning the instance classifier and estimation of the
instance labels.

Deep LLP (DLLP)~\cite{ardehaly2017, dulac2019}, one of the first deep learning
LLP methods, also uses EPRM. DLLP minimizes a~cross-entropy loss on label
proportions, which are obtained by averaging the instance label predictions.
Other deep learning LLP methods build upon DLLP and use different techniques to
improve its robustness. LLP-GAN \cite{liu2019} proposes a generative
adversarial network (GAN) where the discriminator is an instance classifier,
and the generator attempts to generate images similar to the true images.
{LLP-VAT} \cite{tsai2020} adds a~consistency term to the proportion loss, which
serves as a regularization. An advantage of these methods is that they
implicitly support multi-class classification. On the other hand, they require
that bags fit into mini-batches, which might not always be feasible (e.g., for
3D images).

\subsection{Proposed approach}

We focus on the particular case of small bags containing up to about 10
instances. As we will see, this scenario leads to better results by considering
all possible configurations consistent with the annotations explicitly, and
thus avoiding the otherwise necessary approximations, while still
having some interesting applications, such as the embryo classification for
IVF~\cite{hernandez2018, barucic2021}, or analyzing group photos in
Section~\ref{sec:exp}.  

In particular, we model the instance labels as Bernoulli random variables,
whose parameters (i.e., the probabilities of instances to be positive) are
calculated by a~general classification-type neural network from the instance
data.  The number of positive instances in each bag is then a~Poisson binomial
distribution.  We consider two approaches to learn the network's parameters via
maximum likelihood estimation.

First approach (described in Sections \ref{sec:model-and-inference} to
\ref{sec:em-steps}) is a novel EM algorithm based on exact evaluation of the
Poisson binomial distribution.  The inner loop of the EM algorithm consists of
standard supervised training of the underlying network, interleaved with
updating the learning targets using the bag annotations. The iterative and
alternating nature of this algorithm resembles {Alter-$\propto$SVM}. However,
our formulation is based on likelihood maximization, not EPRM, and our
algorithm is more general, allowing to train any instance classifier, namely
a~deep network, which is more powerful than an~SVM.

Second approach is based on approximating the Poisson distribution with an
appropriately parametrized normal distribution.

We experimentally compare the two approaches with two existing LLP methods
based on deep learning, Deep LLP~\cite{ardehaly2017, dulac2019} and
LLP-GAN~\cite{liu2019}. The experiments show that the exact method converges
faster and more reliably to solutions that are comparable to or better than the
competition, while the approximating approach performs comparably to existing
methods, showing the benefit of the exact method.

\section{Methods}

\subsection{Model and inference}
\label{sec:model-and-inference}

We consider a bag $(\bag{X}, y)$, where $\bag{X} = (\inst{x}_1, \inst{x}_2,
\ldots, \inst{x}_n)$ is a set of $n$ instances with unknown binary labels
$h_i\in \{0,1\}$ but known total number of positive labels $y=\sum_{i=1}^n
h_i$. We assume that the sample bag is drawn from a~distribution
\begin{equation}
p(\bag{X}, y) = P_{\vec{\theta}}(y \mid \bag{X}) p(\bag{X}).
\end{equation}
We model the conditional probability $P_{\vec{\theta}}(y \mid \bag{X})$ parametrized by
a vector $\vec{\theta}$;  $p(\bag{X})$ is not modeled as it is not needed for
classification.

Let $\H = \{ \h \in \{0, 1\}^n \mid \sum_{i=1}^n h_i = y \}$ be the
set of all possible instance label configurations $\h$ consistent with the
count $y$. Assuming that the instance label $h_i$ depends only on the instance
data $\inst{x}_i$, the conditional probability $P_{\vec{\theta}}(y \mid \bag{X})$ is
a~Poisson binomial distribution
\begin{align}
P_{\vec{\theta}}(y \mid \bag{X}) &=
\sum_{\h \in \H} P_{\vec{\theta}}(\h \mid \bag{X}) \label{eq:poisson-binom}\\
\text{where}\quad P_{\vec{\theta}}(\h \mid \bag{X}) &= 
\prod_{i=1}^n P_{\vec{\theta}}(h_i \mid \inst{x}_i).
\end{align}

We model the instance-level probabilities $P_{\vec{\theta}}(h_i \mid
\inst{x}_i)$ by a~deep network $f_{\vec{\theta}}$ with weights $\vec{\theta}$
and one sigmoid-activated output,
\begin{equation}
P_{\vec{\theta}}(h_i = 1\mid \inst{x}_i) = f_{\vec{\theta}}(\inst{x}_i) \in [0, 1].
\end{equation}
In this work, $\inst{x}_i$ are images but other input types are possible.

At inference time, the predicted label is determined directly by thresholding
the deep network's output,
\begin{equation}
\widehat{h_i} = 
\left\llbracket f_{\vec{\theta}}(\inst{x}_i) \geq \tau \right\rrbracket,
\quad\text{with}\quad\tau=0.5.
\end{equation}

\subsection{Learning}

We are given $m$ training bags $\mathcal{T} = \{(\bag{X}^j, y^j)\}_{j=1}^m$.
Assuming the independence of the bags, the parameters $\vec{\theta}$ are
estimated by maximizing the estimated conditional log-likelihood 
\begin{equation}
\log P(\mathcal{T} \mid \vec{\theta}) =
\mathcal{L}(\vec{\theta}) 
= \sum_{j=1}^m \log \sum_{\h \in \H^j}
P_{\vec{\theta}}(\h \mid \bag{X}^j).
\end{equation}
We follow the EM approach~\cite{flach2014}, and define
\begin{subequations}
\begin{equation}
\begin{split}
\mathcal{L}(\vec{\theta}, \vec{\alpha}) 
&= \sum_{j=1}^m \sum_{\h \in \H^j} \alpha^j(\h) \log P_{\vec{\theta}}(\h \mid \bag{X}^j) \\
&\qquad\qquad\qquad - \alpha^j(\h) \log \alpha^j(\h)
\end{split}
\end{equation}
\begin{equation}
\text{with} \quad \sum_{\h \in \H^j} \alpha^j(\h) = 1,
\label{eq:alpha-constraints}
\end{equation}
\end{subequations}
where $\alpha^j(\h) \geq 0$ are auxiliary variables corresponding to the (yet
unknown) posterior probabilities $P\bigl(\h \mid  \bag{X}^j\bigr)$.  It
follows from Jensen's inequality that $\mathcal{L}(\vec{\theta}, \vec{\alpha})
\leq \mathcal{L}(\vec{\theta})$.  We shall maximize $\mathcal{L}(\vec{\theta},
\vec{\alpha})$ by alternating the following steps:
\begin{subequations}
\begin{align}
\max_{\vec{\alpha}} &
  \,\sum_{j=1}^m \sum_{\h \in \H^j} 
    \alpha^j(\h) \log P_{\vec{\theta}}(\h \mid \bag{X}^j) - 
    \alpha^j(\h) \log \alpha^j(\h),
\label{eq:e-step-task}\\
\max_{\vec{\theta}} &
    \,\sum_{j=1}^m \sum_{\h \in \H^j} 
    \alpha^j(\h) \log P_{\vec{\theta}}(\h \mid \bag{X}^j).
\label{eq:m-step-task}
\end{align}
\end{subequations}

\subsection{EM steps}
\label{sec:em-steps}

In the \textit{expectation} (E) step, the optimization task
\eqref{eq:e-step-task} is solved w.r.t.  the constraints
\eqref{eq:alpha-constraints}. It can be solved independently for each $j$ using
a~closed-form expression,
\begin{equation}
\alpha^j(\h) = 
\frac{P_{\vec{\theta}}(\h \mid \bag{X}^j)}
     {\sum_{\h' \in \H^j} P_{\vec{\theta}}(\h' \mid \bag{X}^j)},
\quad\text{for}\quad \h \in \H^j.
\end{equation}
The \textit{maximization} (M)~\eqref{eq:m-step-task}
can be rewritten as
\begin{equation}
\max_{\vec{\theta}}
\,\sum_{j=1}^m \sum_{i=1}^{n^j} 
    \phi_i^j \log f_{\vec{\theta}}(\inst{x}_i^j) + 
    (1 - \phi_i^j) \log(1 - f_{\vec{\theta}}(\inst{x}_i^j)),
\end{equation}
which corresponds to the binary cross-entropy loss function, so standard
supervised deep learning techniques can be directly employed.  The coefficients
\begin{equation*}
\phi_i^j = \sum_{\h \in \H^j} \llbracket h_i = 1\rrbracket
\alpha^j(\h)
\end{equation*}
represent the instance probabilities $P_{\vec{\theta}}(h_i^j = 1 \mid
\bag{X}^j, y^j)$. Since updating the $\alpha$ and $\phi$ coefficients is not
computationally demanding, we do it after each epoch. This also makes the
algorithm more stable and robust. We refer to this approach as MLE-LLP.

\subsection{Approximation}

The Poisson binomial distribution \eqref{eq:poisson-binom} becomes intractable
for bigger bags as the number of possible instance label configurations $\lvert
\H^j \rvert$ grows exponentially with the bag size $n^j$. Another
approach \cite{rosenman2018} is to approximate the Poisson binomial
distribution by a normal distribution, $P(y^j \mid \bag{X}^j) \approx
\mathcal{N}(\mu_j, \sigma_j^2)$, where 
\begin{subequations}
\begin{align} 
\mu_j &= \sum_{i=1}^{n^j} f_{\vec{\theta}}(\inst{x}_i^j), \\
\sigma_j^2 &= \sum_{i=1}^{n^j} (1 - f_{\vec{\theta}}(\inst{x}_i^j))f_{\vec{\theta}}(\inst{x}_i^j).
\end{align}
\end{subequations}

Adapting the conditional log-likelihood to this approximation leads to the
following objective
\begin{align}
\min_{\vec{\theta}} \, \sum_{j=1}^m \frac{(y^j - \mu_j)^2}{\sigma_j^2} + \log(\sigma_j^2),
\end{align}
which can be optimized using the standard back-propagation algorithm. We denote
this approach as Approximated MLE-LLP, or AMLE-LLP.

\begin{figure}[t]
\centering
\includegraphics[width=\linewidth]{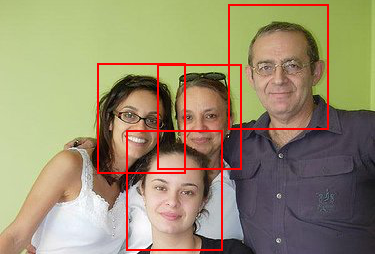}
\caption{Example of an image from which a bag of four instances (framed faces)
is extracted, three of which are positive (female).}
\label{fig:family}
\end{figure}

\begin{figure*}[t]
\centering

\includegraphics{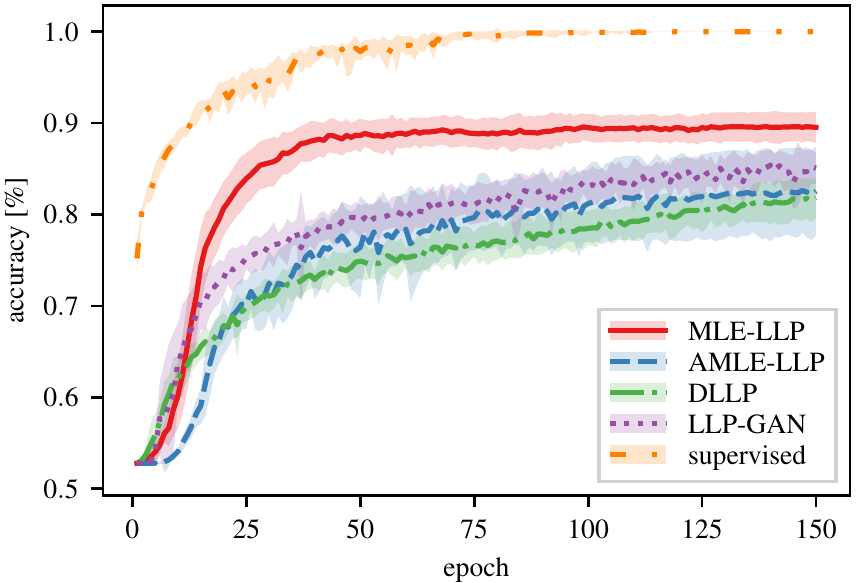}
\hfill
\includegraphics{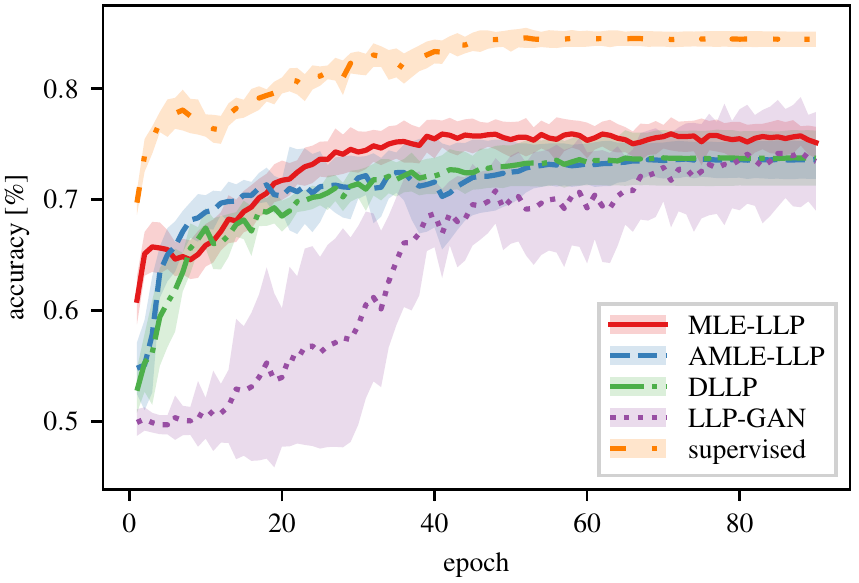}

\caption{Learning curves for the group photos (left) and CIFAR (right)
datasets. The average test accuracy and the standard deviation is plotted using
lines and error bands, respectively. One epoch corresponds to one pass of all
training data through the network.}
\label{fig:learning-curve}
\end{figure*}

\section{Experiments}
\label{sec:exp}

We compare both MLE-LLP and AMLE-LLP with two existing deep LLP methods, DLLP
\cite{ardehaly2017} and LLP-GAN \cite{liu2019}. As a baseline, we use the
standard supervised learning from all instance labels. All LLP methods were
only provided with the number of positive instances in each bag.  All
experiment results were obtained via a 10-fold cross-validation.

The architecture in \cite[Table 2 (CIFAR-10)]{liu2019} was used for LLP-GAN.
The other methods employed the architecture from \cite[Table 3
(CIFAR-10)]{liu2019}. The networks were trained via the Adam \cite{kingma2014}
optimizer.  Image transformations (flipping, blurring, and color adjustments)
were applied to prevent overfitting. The reported results were achieved with
the best hyper-parameters discovered through grid-search for each method
individually.

The experiments were implemented using PyTorch 1.10.0 with TorchVision 0.11.1
and Lightning 1.5.9 and performed on a server equipped with Intel Xeon Silver
4214R (2.40GHz) and NVIDIA GeForce RTX 2080 Ti. 

\subsection{Datasets}

Two different datasets were used in the experiments.  The first dataset, which
deals with sex classification, is based on a~real-world annotated dataset of
family photos~\cite{gallagher2009}.  The annotations provide the position, sex,
and age category of the faces in each image. We extracted the face regions (see
Fig.~\ref{fig:family}), rescaled them to $72 \times 90$ px, and kept only those
of people older than 12 years.  Faces in one image formed a bag. We omitted
images containing more than 12 faces, resulting in ${1\,913}$ bags of
${3\,439}$ and $4\,012$ male and female instances, respectively. Using the LLP
formulation is beneficial in this application, as it is faster for the
annotator to count the number of males/females in the photo than manually
select the respective faces.

The second dataset consisted of the \textit{bird} and \textit{cat} classes from
the CIFAR-10~\cite{cifar10} dataset, a~collection of $32 \times 32$ px color
images. Bags of uniform random sizes $n$, $1 \leq n \leq 12$, were constructed
by uniform random sampling without replacement.

\begin{table}[t]
\centering

\caption{Average epoch duration ($\pm 1\sigma$). For MLE-LLP, the duration includes both the E
and M steps.}
\label{tab:epoch-duration}

\begin{tabular}{@{}lrr@{}}
\toprule
           & \multicolumn{2}{c}{epoch duration [s]} \\
\cmidrule(l){2-3}
method     &   group photos   &     CIFAR         \\ 
\midrule
MLE-LLP    & $16.66 \pm 0.25$ & $ 7.78 \pm 0.16$  \\
AMLE-LLP   & $29.33 \pm 0.33$ & $20.23 \pm 0.20$  \\
DLLP       & $30.38 \pm 0.31$ & $18.66 \pm 0.19$  \\
LLP-GAN    & $77.01 \pm 0.78$ & $51.20 \pm 0.29$  \\
supervised & $12.81 \pm 0.19$ & $ 3.65 \pm 0.13$  \\
\bottomrule
\end{tabular}
\end{table}

\subsection{Convergence}

The first experiment focused on the convergence of the methods on both
datasets. The average epoch time was measured, and the learning curves were
plotted for all methods.

For both datasets, the supervised method converged to an accuracy that was
$\approx 10\%$ higher than the LLP methods (see Fig.~\ref{fig:learning-curve}).
On the group photos, MLE-LLP led steadily to a~better performance than the
competitors, while on CIFAR, the methods converged to a comparable accuracy.
Furthermore, MLE-LLP exhibited lower variance of its performance than the other
LLP methods.

Among the LLP methods, MLE-LLP was the most efficient in terms of time as it
required significantly fewer epochs for convergence than the other LLP methods,
including AMLE-LLP: approximately $3 \times$ fewer for the group photos and $2
\times$ for CIFAR. Moreover, one epoch of MLE-LLP was about twice as fast on
average than for the other LLP methods (see Tab.~\ref{tab:epoch-duration}). The
speed was achieved mainly by processing the data in bigger batches, which
MLE-LLP natively enables.

\subsection{Bag size impact}

We considered seven bag sizes $n \in \{ 2^k \mid k = 1, \ldots, 7\}$. For each
bag size, we generated a dataset by randomly sampling instances from CIFAR-10
without replacement, creating bags of size $n$. MLE-LLP was tested only for small bag sizes where its computation requirements were reasonable.

Generally, the methods behave similarly. Their performance degrades as the
bag size grows: the accuracy decreases by $\approx 5\%$ each time the bag size
doubles. This follows from the fact that, as the bag size grows, the
distribution of instance labels in a bag approaches their a~priori
distribution, which is uniform in this case, limiting the information available to LLP~\cite{yu2015}.

\begin{figure}[t]
\centering
\includegraphics[width=\linewidth]{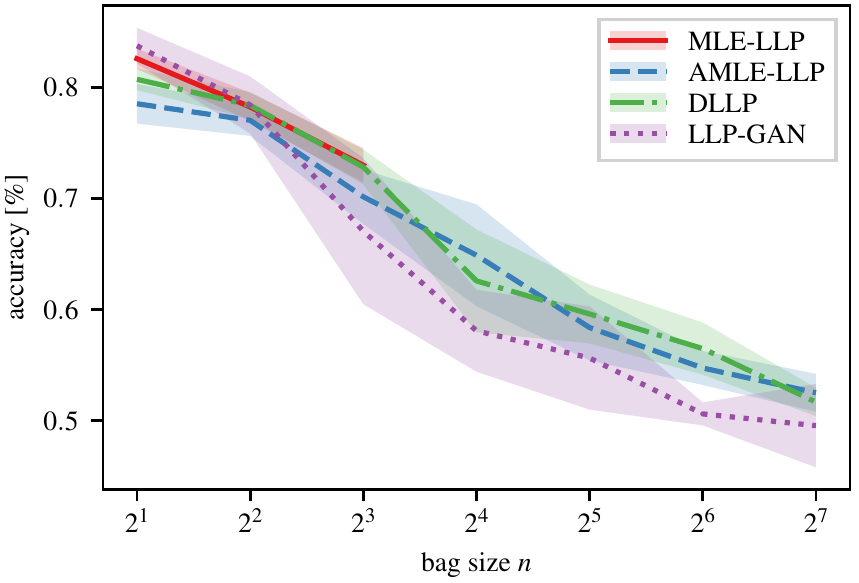}

\caption{Dependency of the performance of the LLP methods on the bag size.}
\label{fig:bagsize}
\end{figure}

\section{Conclusion}

This paper considered learning from label proportions (LLP) when the training
dataset consists of small bags.  Our MLE-LLP method is general, it can be used
with any instance classifier.  It considers all possible configurations
consistent with the annotations, which allows for faster convergence than
existing methods based on deep learning.  Faster convergence is especially
desirable when big datasets consisting of high-dimensional instances are
processed. Unlike existing methods, MLE-LLP enables processing the data in
arbitrarily-sized batches. The source code and the data is available online\footnote{\url{https://github.com/barucden/mlellp}}.

\subsection*{Acknowledgments}

{\small The authors acknowledge the support of the OP VVV funded project ``CZ.02.1.01/0.0/0.0/16\_019/0000765 Research Center for Informatics'', the Czech Science Foundation project 20-08452S, and the Grant Agency of the CTU in Prague, grant No. SGS20/170/OHK3/3T/13. \par}

\bibliographystyle{IEEEbib}
\balance
\bibliography{references}

\end{document}